\documentclass[pdflatex,sn-mathphys]{sn-jnl}

\jyear{2021}%

\theoremstyle{thmstyleone}%
\theoremstyle{thmstyletwo}%

\theoremstyle{thmstylethree}%

\usepackage{derivative}
\usepackage[autostyle=false, style=english]{csquotes}
\MakeOuterQuote{"}
\usepackage{xr}
\makeatletter

\newcommand*{\addFileDependency}[1]{
\typeout{(#1)}
\@addtofilelist{#1}
\IfFileExists{#1}{}{\typeout{No file #1.}}
}\makeatother

\newcommand*{\myexternaldocument}[1]{%
\externaldocument{#1}%
\addFileDependency{#1.tex}%
\addFileDependency{#1.aux}%
}

\myexternaldocument{SI}

\usepackage{lineno}
\usepackage{booktabs}
\usepackage{multirow}
\usepackage{graphicx} 
\usepackage{subcaption}
\usepackage[section]{placeins}
\usepackage{tabularx}
\usepackage{multirow}
\usepackage{CJKutf8}
\usepackage{graphicx} 
\graphicspath{{figures/}} 
\usepackage{array} 
\usepackage{makecell} 
\usepackage[export]{adjustbox}

\usepackage{float}
\usepackage{etoolbox}
\makeatletter
\patchcmd{\ps@headings}
{\hbox to \hsize{\hfill Springer Nature 2021 \LaTeX\ template\hfill}}
{\hbox to \hsize{}}
{}
{}
\patchcmd{\ps@titlepage}
{\hbox to \hsize{\hfill Springer Nature 2021 \LaTeX\ template\hfill}}
{\hbox to \hsize{}}
{}
{}
\makeatother
\makeatletter
\patchcmd{\ps@headings}
{\hbox to \hsize{\hfill Springer Nature 2021 \LaTeX\ template\hfill}}
{\hbox to \hsize{}}
{}
{}
\patchcmd{\ps@headings}
{\hbox to \hsize{\hfill Springer Nature 2021 \LaTeX\ template\hfill}}
{\hbox to \hsize{}}
{}
{}
\patchcmd{\ps@titlepage}
{\hbox to \hsize{\hfill Springer Nature 2021 \LaTeX\ template\hfill}}
{\hbox to \hsize{}}
{}
{}
\makeatother

\begin{document}

\begin{CJK*}{UTF8}{gbsn}

\title[FuXi-Uni]{A unified multimodal understanding and generation model for cross-disciplinary scientific research}

\author[2]{\fnm{Xiaomeng} \sur{Yang}}\email{yangxiaomeng@sais.org.cn}
\equalcont{These authors contributed equally to this work.}

\author[1,2]{\fnm{Zhiyu} \sur{Tan}}\email{zytan24@m.fudan.edu.cn}
\equalcont{These authors contributed equally to this work.}

\author[1,2,4,5]{\fnm{Xiaohui} \sur{Zhong}}\email{x7zhong@gmail.com}
\equalcont{These authors contributed equally to this work.}

\author[1,2]{\fnm{Mengping} \sur{Yang}}\email{yangmengping@sais.org.cn}

\author[1,2,3]{\fnm{Qiusheng} \sur{Huang}}\email{qiusheng.shawn@gmail.com}

\author[1,2]{\fnm{Lei} \sur{Chen}}\email{cltpys@163.com}

\author*[6,2,3,7]{\fnm{Libo} \sur{Wu}}\email{wulibo@fudan.edu.cn}

\author*[1,2,3,5]{\fnm{Hao} \sur{Li}}\email{lihao$\_$lh@fudan.edu.cn}

\affil[1]{\orgdiv{Artificial Intelligence Innovation and Incubation Institute}, \orgname{Fudan University}, \orgaddress{\city{Shanghai}, \postcode{200433}, \country{China}}}

\affil[2]{\orgname{Shanghai Academy of Artificial Intelligence for Science}, \orgaddress{\city{Shanghai}, \postcode{200232}, \country{China}}}

\affil[3]{\orgname{Shanghai Innovation Institute}, \orgaddress{\city{Shanghai}, \postcode{200231}, \country{China}}}

\affil[4]{\orgname{FuXi Intelligent Computing Technology Co., Ltd.}, \orgaddress{\city{Shanghai}, \postcode{200233}, \country{China}}}

\affil[5]{\orgname{Joint Laboratory for AI-Based Earth System Forecasting}, \orgaddress{\city{Shanghai}, \country{China}}}

\affil[6]{\orgdiv{School of Data Science}, \orgname{Fudan University}, \orgaddress{\city{Shanghai}, \postcode{200433}, \country{China}}}


\affil[7]{\orgdiv{MOE Laboratory for National Development and Intelligent Governance}, \orgname{Fudan University}, \orgaddress{\city{Shanghai}, \postcode{200433}, \country{China}}}


\abstract{
Scientific discovery increasingly relies on integrating heterogeneous, high-dimensional data across disciplines nowadays.
While data-driven artificial intelligence (AI) models have achieved notable success across various scientific domains, they typically remain domain-specific or lack the capability of simultaneously understanding and generating multimodal scientific data, particularly for high-dimensional data.
Yet, many pressing global challenges and scientific problems are inherently cross-disciplinary and require coordinated progress across multiple fields.
Here, we present FuXi-Uni, a native unified multimodal model for scientific understanding and high-fidelity generation across diverse scientific domains within a single architecture.
Specifically, FuXi-Uni aligns cross-disciplinary scientific tokens within natural language tokens and employs science decoders to reconstruct scientific tokens, thereby supporting both natural language conversation and scientific numerical prediction.
Empirically, we validate FuXi-Uni in Earth science and Biomedicine.
In Earth system modeling, the model supports global weather forecasting, tropical cyclone (TC) forecast editing, and spatial downscaling driven by only language instructions. 
FuXi-Uni generates 10-day global forecasts at 0.25$^\circ$ resolution that outperform the state-of-the-art (SOTA) physical forecasting system.
It shows superior performance for both TC track and intensity prediction relative to the SOTA physical model, and generates high-resolution regional weather fields that surpass standard interpolation baselines.
Regarding biomedicine, FuXi-Uni outperforms leading multimodal large language models (MLLMs) on multiple biomedical visual question answering benchmarks.
By unifying heterogeneous scientific modalities within a native shared latent space while maintaining strong domain-specific performance, FuXi-Uni provides a step forward more general-purpose, multimodal scientific models.
This approach suggests a scalable foundation for integrated, AI-assisted scientific research and solutions to global challenges.
}

\keywords{multimodal, understanding, generation, science}

\maketitle

\section{Introduction}

Traditional scientific discovery is fundamentally a data-driven process, with scientists and researchers formulating and validating theories based on observational data from natural phenomena and reproducible experiments \cite{hey2009fourth,fortunato2018science,AI4SFDU2025}.
Scientific data are inherently heterogeneous and exist in diverse modalities, including electronic health records, medical imaging, biosensor measurements in biomedicine \cite{acosta2022multimodal,liu2025challenges}; weather station records, satellite imagery, and numerical simulations in Earth science\cite{Li2022,Tiffany2024,Wang_GeoGPT_2025}; and spectroscopic characterization, atomic structures, microscopic images, and text-based synthesis protocols in materials science \cite{gong2023multimodal,ock2024unimat}.
Rapid advances in high-throughput experimental platforms, increased sensor density and resolution, and finer-grained computational modeling have triggered an explosive growth in the volume, variety and complexity of scientific data \cite{hey2009fourth,stephens2015big,Li2023BigDataEarthSystem,Yang2024BiomedicalBigData,Tiffany2024,Zafar2025}.
This data deluge presents unprecedented opportunities but also overwhelms conventional analytical approaches.
Meanwhile, artificial intelligence (AI) has emerged as a powerful new paradigm for scientific discovery due to its capability to extract structure and actionable insights from large-scale, multimodal datasets \cite{Reichstein2019,wang2023scientific,Yannis2024,AI4SFDU2025,Nina2025}.

The past few years have witnessed remarkable breakthroughs in AI for Science, wit AI evolving into a constellation of domain-specific foundation models that exploit the intrinsic structure of scientific data and achieve previously unimaginable performance.
This transformation reached a historic milestone in 2024, when the Nobel Prize in Physics recognized foundational contributions to the development of artificial neural networks, and the Nobel Prize in Chemistry honored computational protein design and protein structure prediction, explicitly acknowledging AI-driven scientific discovery as a new pillar of the natural sciences \cite{li2024artificial,hopfieldai}.
In the life sciences, the frontier has advanced beyond single-protein structure prediction with AlphaFold \cite{Jumper_Highly_accurate_2021}, toward unified prediction of biomolecular complexes as exemplified by AlphaFold 3 \cite{abramson2024accurate}.
In Earth science, spatio-temporal foundation models trained on reanalysis dataset \cite{hersbach2020era5} now surpass conventional numerical weather prediction (NWP) models \cite{pathak2022fourcastnet,bi2023accurate,lam2022graphcast,chen2023fuxi,bouallegue2024aifs,chen2024machine,kochkov2024neural,price2025probabilistic,bodnar2025foundation,fuxiens2025} in forecast accuracy while offering substantial advantages in computational efficiency.
In materials science, AI is accelerating property prediction, enabling inverse design, and seamlessly integrating with automated experimentation to transform materials discovery and engineering, significantly exceeding the pace of traditional high-throughput computation or manual experimentation \cite{butler2018machine,schuh2022accelerating,merchant2023scaling,pyzer2025foundation,zeni2025generative}.
Despite these successes, most existing models remain limited to single domains, restricting their ability to address interdisciplinary challenges that require a unified understanding across scientific domains.

However, many of today's most pressing global challenges, including climate change and climate tipping points \cite{kay2015community,lenton2019,masson2021climate,Luke2022}, robotics and autonomous systems \cite{Koditschek2021,ribeiro2021robotic,guenat2022meeting}, pandemic and epidemic intelligence \cite{morgan2022better,williams2023outlook}, sustainable energy transitions \cite{Rogelj2015Paris,Bistline2021NetZero,zylstra2024targetGain,kruse2025predictingIgnition}, and safe AI, are inherently complex and cross-disciplinary.
Addressing these challenges requires integrating expertise from Earth science, life science, biomedicine, material science, social science, and AI within a unified framework.
Yet mastering the necessary breadth and depth of knowledge typically takes years to decades of training for individual researchers.
Therefore, unified multimodal models capable of learning across scientific domains and data modalities offer a promising alternative to accelerate coherent, systemic solutions.
Instead of having separate models for each task or modality, a unified multimodal understanding and generation models is a single AI model that processes multiple data types (e.g., text, images, audio, and video) and generate outputs in one or more modalities within a single architecture and shared token space, enabling joint optimization of understanding and generation across multiple domains \cite{zhang2025unified}.
The release of GPT-4o \cite{GPT4O} in March 2025 further demonstrated the potential of such unified multimodal architectures, showing that joint training for understanding and generation can mutually reinforce both capabilities.
GPT-4o exhibits improved performance in generating images following complex instructions, reasoning over visual inputs, and producing coherent multimodal analyses across synthesized outputs.



Several recent studies \cite{hu2025surveyscientificLLM} transform cross-disciplinary scientific data, including diverse scientific data types along with scientific text, into discrete representations to align them with language features used by large language models (LLMs) \cite{li2025discretetokenization}.
Most of these approaches build on general-purpose LLMs, either by through pre-training or finetuning them on data from specific scientific tasks \cite{xie2023darwin,eger2025transforming} or by pre-training directly on scientific data to enhance performance on scientific applications \cite{Galactica2022,SciDFM2024,OmniScience2025}.
While this strategy leverages LLMs’ knowledge, reasoning capabilities, and ecosystem, mapping high-dimensional fields, geometric structures, and dynamical systems into one-dimensional tokens inevitably introduces information loss and can degrade domain-specific tasks such as quantitative prediction and decision support.
This limitation is particularly severe in scientific domains with extremely high-dimensional and complex data.
For instance, in global weather prediction at spatial resolution of 0.25$^\circ$, a single snapshot can have dimensionality, $\textrm{C}\times721\times1440$, where $\textrm{C}\ge65$ corresponds to 13 pressure levels times 5 atmospheric variables to represent the three-dimensional (3D) atmospheric structure, and 721 and 1440 are the numbers of grid points in latitude and longitude, respectively \cite{pathak2022fourcastnet,bi2023accurate,lam2022graphcast,chen2023fuxi,bouallegue2024aifs,chen2024machine,kochkov2024neural,price2025probabilistic,bodnar2025foundation,fuxiens2025}.
Similarly, high-energy physics experiments such as A Toroidal LHC Apparatus (ATLAS) and Compact Muon Solenoid (CMS) at the Large Hadron Collider generate collision data at rates of tens of terabytes per second \cite{evans2008lhc}.
Consequently, current cross-disciplinary scientific LLMs are largely limited to understanding multimodal scientific data, and cannot support more complex generative tasks such as high-fidelity weather prediction.


To enable cross-disciplinary understanding and generation while maintaining strong domain-specific performance, we introduce FuXi-Uni, a prototype framework that unifies scientific and textual modalities within a shared latent space and supports a wide range of scientific tasks via natural language instructions.
Unlike text-centric paradigms, FuXi-Uni employs domain-aware science tokenizers designed for extremely high-dimensional scientific data, preserving complex field structures and spatiotemporal dynamics.
This design, for the first time, aligns heterogeneous, high-dimensional scientific data with text, mitigates information loss from discretization, and enables generation of both textual outputs and high-dimensional numerical outputs.
To the best of our knowledge, FuXi-Uni is the first AI model to achieve unified multimodal understanding and generation across multiple scientific domains.
We validate FuXi-Uni in Earth science and biomedicine, where it achieves state-of-the-art (SOTA) performance on the following tasks:
\begin{itemize}
  \item the first AI model to simultaneously perform weather prediction, bias correction, and spatial downscaling, outperforming the SOTA NWP model;
  \item 10-day global weather forecasts at 0.25$^{\circ}$ spatial resolution and 6-hourly temporal resolution, outperforming the SOTA NWP model, the European Centre for Medium-Range Weather Forecasts (ECMWF) high-resolution forecast (HRES) \cite{ECMWF2021};
  \item editing tropical cyclone (TC) forecasts via prompting, enabling AI-based weather prediction models to outperform conventional NWP models (i.e., ECMWF HRES) in both TC track and intensity for the first time (previous AI models surpassed ECMWF HRES only in track forecasts while underestimating TC intensity);
  \item spatial downscaling from 1.5$^{\circ}$ to 0.25$^{\circ}$ resolution, with downscaled forecasts outperforming linear interpolation in both accuracy and image quality;
  \item biomedical visual question answering (VQA), outperforming previous representative SOTA multimodal LLMs on several metrics across three standard biomedical VQA datasets.
\end{itemize}
By aligning scientific domains with LLM-based representations, FuXi-Uni supports a broad range of functionalities.
Through learning from multidisciplinary data with a shared backbone, FuXi-Uni has the potential to move beyond domain-specific foundation models, facilitating more efficient and seamless transfer of expertise across diverse scientific tasks and supporting solutions to global challenges.

\section{FuXi-Uni}

Current unified multimodal understanding and generation models generally follow two architectural paradigms: autoregressive models and hybrid models, reflecting different trade-offs in understanding, generation, and system complexity.
Autoregressive models treat all modalities as discrete token sequences, discretizing heterogeneous data into a shared vocabulary for unified modeling \cite{sun2024emu2,wang2024emu3,wu2024janus,chameleonteam2025}.
A single transformer is then employed to perform next-token prediction over this vocabulary, facilitating unified understanding and generation by sequence modeling.
Despite their conceptual elegance, autoregressive models shows slow inference and excessive token costs when discretizing high-dimensional data such as images or videos.
To overcome these limitations, hybrid models integrate autoregressive components for language understanding with diffusion or other high-fidelity generative models for visual generation \cite{Transfusion2024,show-o2025,deng2025BAGEL,JanusFlow2025}.
Although hybrid models achieve SOTA performance in both understanding and generation, they introduce increased architectural complexity, optimization challenges, and greater inference latency.

Irrespective of paradigm differences, unified models typically consists of three major components: a modality-specific encoder that project raw heterogeneous inputs into a shared representation space for alignment and processing, a modality-fusion backbone that integrates encoded features across modalities, facilitating cross-modal interactions, and a modality-specific decoder that produce outputs in the target modality through autoregressive or denosing processes \cite{zhang2025unified}.
However, extending these architectures to high-dimensional scientific data, (e.g., 3D global atmospheric fields, high-resolution time series, or multi-spectral tensors) remains challenging.
Standard discrete tokenization schemes are particularly inefficient for continuous spatiotemporal data as they induce severe token explosion, produce prohibitive long token sequences, and cause substantial information loss.
Consequently, such approaches fail to preserve the fine-grained physical structures and correlations fundamental to scientifically accurate modeling \cite{nguyen2023climax,Zhang2025}.

\begin{figure}[htbp]
  \centering
  \includegraphics[width=\linewidth]{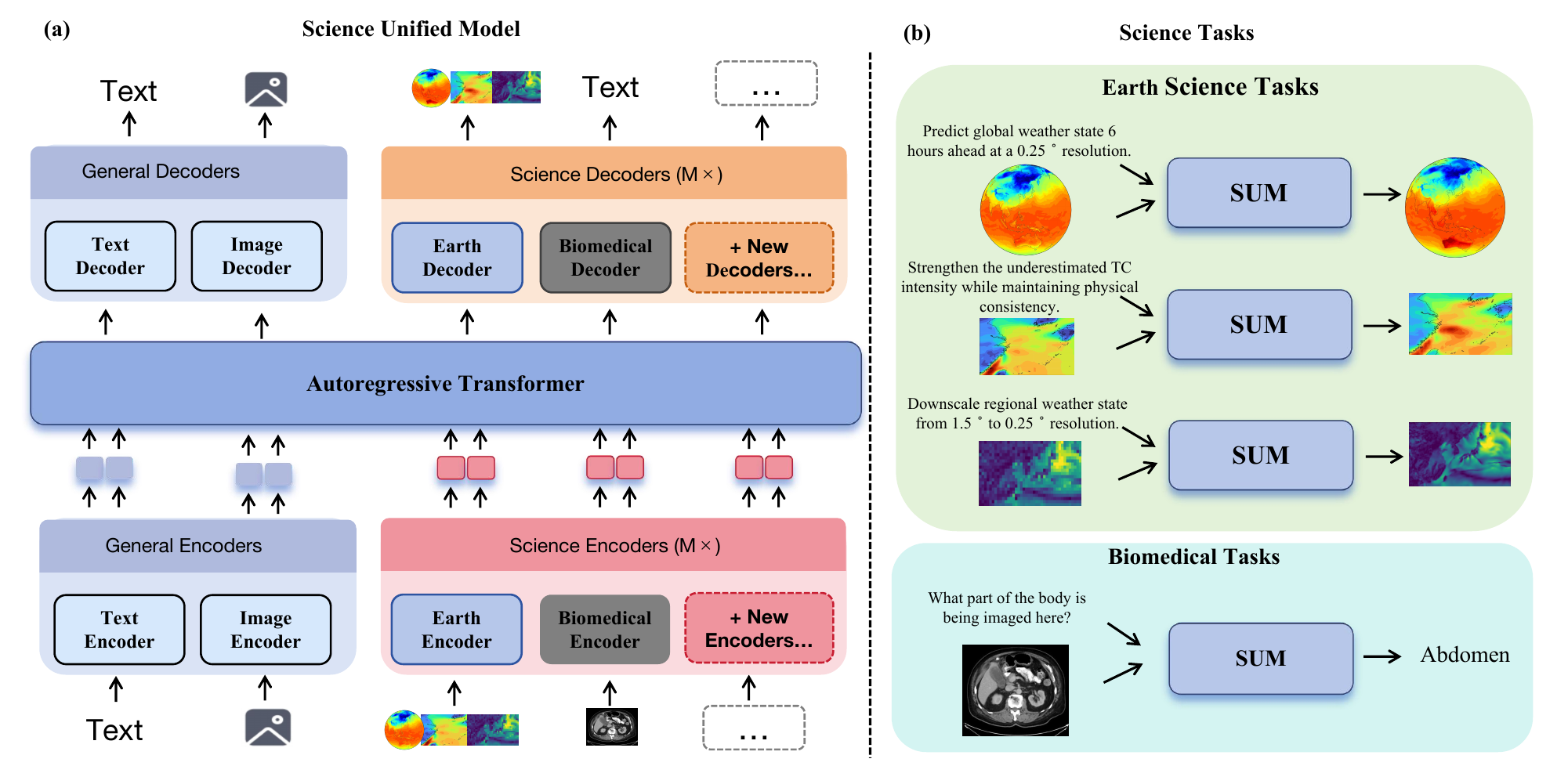}
  \caption{\textbf{Science Unified Model (SUM) overview and applications.}
\textbf{(a)} SUM builds on a shared autoregressive Transformer backbone and general multimodal components with plug-and-play \emph{science} encoders/decoders ($M{\times}$), enabling seamless extension to new scientific modalities and domains.
\textbf{(b)} Representative tasks supported by SUM, including global weather forecasting, physically consistent enhancement of underestimated tropical-cyclone intensity, regional weather downscaling, and biomedical image understanding.}
\label{model_framework}
\end{figure}

To advance beyond these constraints, we present FuXi-Uni, a science-token aligned LLM designed for unified multimodal scientific understanding and generation.
FuXi-Uni directly addresses the limitations of patch-based or discretization tokenization for scientific data by employing domain-specific encoders that transform raw scientific data into structured tokens preserving native spatiotemporal organization.
Building on the design principles of general unified models \cite{zhang2025unified}, FuXi-Uni encompasses (i) domain-specific encoders serving as scientific tokenizers, (ii) a modality-fusion backbone for aligning and integrating multi-domain token representations within a unified latent space, and (iii) scientific task decoders tailored to domain-specific output generation (Figure \ref{model_framework}).
While adopting a hybrid architecture, FuXi-Uni replaces diffusion-based decoders with scientific domain-specific decoders, substantially reducing inference latency.
Text is jointly aligned with scientific tokens within the shared backbone, enabling unified natural language generation and accurate scientific modeling.

FuXi-Uni is built upon the pretrained Qwen2.5-VL-7B vision-language model \cite{bai2025qwen2}.
Its language backbone is a decoder-only Transformer (Qwen2.5-7B) that incorporates rotary positional embeddings (RoPE) \cite{su2024roformer}, RMSNorm \cite{zhang2019root}, and SwiGLU feed-forward blocks \cite{shazeer2020glu}, and grouped-query attention (GQA) \cite{ainslie2023gqa} for enhanced computational efficiency.
For vision-language inputs, FuXi-Uni retains the native Qwen2.5-VL visual pathway, which employs a dynamic-resolution Vision Transformer (ViT) with windowed attention to encode images into visual embeddings, followed by a multi-layer perceptron (MLP)-based merger that compresses these features prior to multimodal fusion.
Building on this architecture, we introduce an Earth-science encoder that maps gridded fields into multimodal tokens compatible with the Qwen2.5-VL backbone; these tokens are co-attended with textual tokens within the shared transformer to enable multimodal fusion across domains.

In the Earth science domain, the input is a four-dimensional data cube, $\textbf{X} \in \mathbb{R} ^{\textrm{T} \times \textrm{C} \times \textrm{H} \times \textrm{W}} $, representing a multivariate weather state at a single time step.
The temporal dimension is fixed at $\textrm{T} = 1$, denoting either the immediately preceding step for weather forecasting, or the same step as the target for TC editing and spatial downscaling.
The channel dimension includes $\textrm{C} = 70$ physical variables (Table \ref{glossary_weather}), while ${\textrm{H}}$ and ${\textrm{W}}$ represent the numbers of grid points along latitude and longitude, respectively.
The spatial dimensions ($\textrm{H}, \textrm{W}$) depend on the target task.
For global weather forecasting, the model ingests global fields of size $\textrm{H} = 721$ and $\textrm{W} = 1440$, corresponding to a spatial resolution of 0.25$^\circ$.
TC editing targets a regional domain (Figure \ref{TC_edit}) at the same resolution, with ${\textrm{H}} = 201$ and ${\textrm{W}} = 240$.
Spatial downscaling uses a coarser regional input (Figure \ref{spatial_downscaling}) of ${\textrm{H}} = 20$ and ${\textrm{W}} = 40$, representing 1.5$^\circ$ resolution.
Despite their differences, all three tasks share a consistent data representation.

To integrate these heterogeneous tasks within a unified modeling framework, the model is conditioned on task-specific textual prompts that specify the intended operation, while scientific tokens represent the underlying weather state.
The prompts used are: \emph{"Predict global weather state 6 hours ahead at a 0.25$^\circ$ resolution"} for global weather forecasting, \emph{"Strengthen the underestimated TC intensity while maintaining physical consistency"} for TC editing, and \emph{"Downscale regional weather state from 1.5$^\circ$ to 0.25$^\circ$ resolution"} for spatial downscaling (Figure \ref{model_framework}).
This prompt-based conditioning provides a consistent interface that allows a single model architecture to flexibly accommodate diverse Earth science tasks.

Beyond Earth science, the same unified framework extends naturally to biomedical vision–language tasks, demonstrating its cross-domain applicability and scalability.
In biomedicine, each sample consists of a two-dimensional (2D) medical image paired with a natural language question, forming a standard biomedical VQA instance across imaging modalities, including X-Ray, computed tomography (CT), magnetic resonance imaging (MRI), microscopy, dermoscopy, and pathology.
We adopt the Qwen2.5-VL VQA architecture and its vision preprocessing pipeline to ensure consistency between heterogeneous inputs.
Arbitrary-resolution images are rescaled to the nearest multiple of 28 and patchified into variable-length visual token sequences, which are processed with window attention to preserve computational efficiency.
The resulting tokens are subsequently compressed via $2\times2$ token merging and an MLP projection before fusion with the LLM, while dynamic resolution sampling combined with absolute coordinate embeddings further enhances cross-resolution generalization and yields robust performance across biomedical imaging scales.

To support training across heterogeneous biomedical benchmarks within this unified architecture, we employ instruction-based supervision, thereby aligning dataset-specific requirements under a shared language interface.
Benchmark-specific prompts are designed for VQA-RAD \cite{VQA_RAD}, SLAKE \cite{SLAKE}, and PathVQA \cite{PathVQA} to enforce dataset-dependent answer formats, including as yes/no, and concise descriptive responses.
For instance, open-ended PathVQA questions use the instruction: \emph{"You are a pathology VQA assistant. Answer the question in a very concise way using a short medical phrase or at most one short sentence. Do not explain your reasoning or add extra text."}).
These structured instructions allow the training splits of all three benchmarks to be merged into a unified instruction mixture for pretraining, enabling a single checkpoint to generalize across all three testing datasets while mitigating catastrophic forgetting relative to sequential per-benchmark fine-tuning.

In summary, FuXi-Uni framework generalizes beyond individual scientific domains and tasks, and thus offers a scalable AI foundation model for interdisciplinary scientific research.
Looking ahead, the framework can be extended from both data and model perspectives. 
From the data perspective, broader domain coverage can be achieved by incorporating additional domain-specific datasets paired with text (e.g., descriptions, question–answer pairs, or procedural narratives), allowing the model to acquire new scientific concepts and task semantics through a unified, prompt interface. 
From the model perspective, extension involves introducing a new scientific-domain encoder that maps the additional scientific modalities into structured scientific tokens compatible with the shared LLM, optionally augmented with a lightweight, domain-specific decoder for target tasks.
Together, these extensions point towards a unified scientific modeling paradigm capable of bridging different scientific domains within a single unified multimodal scientific model.

\section{Results}



\subsection{Earth system modeling}

Accurate weather forecasts are essential for timely decision-making in weather‑sensitive sectors, including agriculture, renewable energy, transportation, and disaster risk management and reduction, yielding substantial socio‑economic benefits and reducing losses \cite{katz1997economic,koetse2009impact,lesk2016influence,hoeppe2016trends,sweeney2020future,wang2023accelerating}.
Since the mid-twentieth century, operational forecasting has primarily relied on NWP models, widely regarded as one of the greatest scientific achievements of the twentieth century.
The importance of NWP development was further recognized by the 2021 Nobel Prize in Physics, awarded to Syukuro Manabe \cite{harper200750th,bauer2015quiet,Nobel2021,Ravishankara2022}.
Despite their success, NWP models are computationally expensive and limited in their ability to represent physical processes on unresolved spatial scales \cite{stensrud2009parameterization}.
These limitations have motivated the development of AI-based forecasting models that match or surpass conventional NWP performance while being orders of magnitude more computationally efficient \cite{bouallegue2024rise,zhang2025machine,Waqas2025}.

Nevertheless, operational forecasting workflows do not end with the generation of raw NWP or AI model outputs.
Weather centers routinely apply post-processing, through human forecaster expertise and statistical or AI-based approaches to correct biases, calibrate forecast uncertainty, and tailor forecasts to user-relevant variables and locations \cite{inness2012operational,gneiting2014calibration}.
Such post-processing is particularly critical for forecast products such as TC forecasts, for which both conventional NWP and current AI models exhibit substantial underestimations in TC intensity prediction \cite{bi2023accurate,lam2022graphcast,chen2023fuxi,xu2025exploring,niu2025improving,guo2025fuxitc}.
Additionally, many applications require spatial downscaling to recover high-resolution details from coarse-resolution model output, particularly in regions with complex terrain \cite{downscaling2024,Rampal2024,hess2025fast}.
This task is closely analogous to super-resolution in computer vision \cite{Dong2016,Vandal2017,Vandal2019}.
FuXi-Uni leverages the capabilities of LLMs to interactively integrate human forecaster expertise with AI-based correction and downscaling models within a unified framework that can work as an "AI meteorological forecaster".
To our knowledge, it is the first AI model capable of simultaneously performing weather prediction, bias correction, and spatial downscaling according to text-based instructions, which also substantially simplifies deployment and maintenance by reducing reliance on multiple task-specific models.



This subsection evaluates FuXi-Uni's performance in generating 10-day global weather forecasts at 0.25$^{\circ}$ spatial and 6-hour temporal resolution, performing prompt-guided TC forecast correction, and conducting spatial downscaling from 1.5$^{\circ}$ to 0.25$^{\circ}$ resolution.
FuXi-Uni's global forecasting skill is compared with that of ECMWF HRES.
Following standard practice \cite{rasp2020weatherbench,rasp2024weatherbench}, FuXi-Uni and ECMWF HRES are evaluated against ERA5 reanalysis and the 0-hour lead time analysis of HRES (HRES-fc0), respectively (see Section \ref{evaluation_method}).
TC forecasts are assessed using the International Best Track Archive for Climate Stewardship (IBTrACS) \cite{knapp2010,gahtan2024international} as the reference.

\subsubsection{Global weather forecasting}


This subsection compares the 6-hourly global weather forecast performance of FuXi-Uni and ECMWF HRES a function of forecast lead time up to 10 days.
In evaluating global weather forecasts, we focus on 500 hPa geopotential height (Z500), 2-m temperature (T2M), and 10-m wind speed (WS10M), which characterize large-scale mid-tropospheric circulation, near-surface thermodynamic conditions, and boundary-layer dynamical processes, respectively.
Together, these variables provide a physically meaningful and widely adopted basis for assessing the skill of global numerical and AI-based weather prediction models \cite{Hansen2006,Zheng2007,McVicar2012,Christidis2015}.

\begin{figure}[h]
\centering
\includegraphics[width=\linewidth]{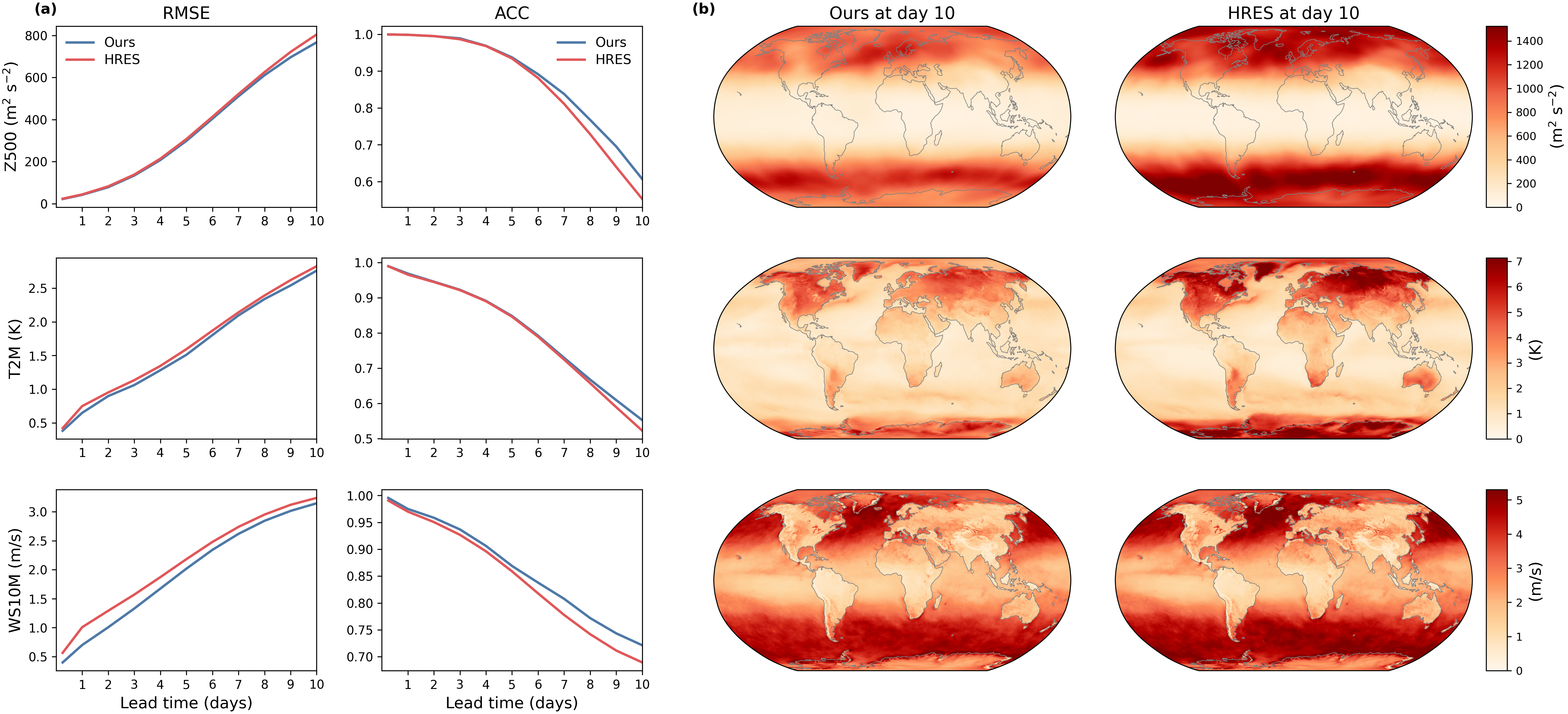}
\caption{\textbf{FuXi-Uni outperforms ECMWF HRES in 10-day weather forecasts.} \textbf{a,} Comparison of globally-averaged and latitude-weighted root mean square error ($\textrm{RMSE}$, first column) and anomaly correlation coefficient (ACC, second column) for weather forecasts from FuXi-Uni (blue lines) and ECMWF HRES (red lines). \textbf{b,} Spatial distributions of average $\textrm{RMSE}$ without latitude weighting of forecasts from FuXi-Uni (first column) and ECMWF HRES (second column) forecasts at day 10. Results are shown for three variables: 500 hPa geopotential ($\textrm{Z500}$, first row), 2-meter temperature ($\textrm{T2M}$, second row), and 10-meter wind speed ($\textrm{WS10M}$, third row), calculated using all testing data over a 1-year testing period (June 01, 2023 - June 30, 2024).}
\label{global_weather_forecast}
\end{figure}

Figure \ref{global_weather_forecast}a presents the globally-averaged and latitude-weighted root mean square error (RMSE) and anomaly correlation coefficient (ACC) \cite{murphy1989skill,hamill2013noaa} as a function of forecast lead times.
RMSE measures the mean magnitude of forecast errors, while ACC quantifies the correlation between predicted and observed anomalies and reflects the ability to capture large-scale synoptic patterns.
Lower RMSE and higher ACC values suggest better performance.
Details of the metric calculations are provided in Section \ref{evaluation_method}.

FuXi-Uni consistently outperforms ECMWF HRES, exhibiting lower RMSE and higher ACC across all variables throughout the 10-day forecasts.
Notably, although all previous AI-based models \cite{bi2023accurate,lam2022graphcast,chen2023fuxi} require meteorological inputs from two preceding time steps, FuXi-Uni still mange to achieve superior forecasting performance relative to ECMWF HRES using only a single single time step.
This design enables a unified framework consistent with other Earth science tasks, such as TC editing and spatial downscaling, which operate on a single input data cube.
Figure \ref{global_weather_forecast}b further shows the spatial distribution of the average $\textrm{RMSE}$ without latitude weighting for forecasts from FuXi-Uni and ECMWF HRES at day 10.
Darker red shades indicate higher RMSE values, showing errors increasing from the tropics to extra-tropical regions for both models.
FuXi-Uni shows systematically lower RMSE, particularly in extratropics, indicating superior long-range forecast skill.



\FloatBarrier

\subsubsection{Tropical cyclone forecasts editing}
TCs are among the most devastating and costly natural disasters globally \cite{shultz2005epidemiology,krichene2023social}, causing substantial loss of life and socio‑economic damage.
Therefore, improving TC forecasts is critical for effective disaster preparedness and response.
While recent AI‑based weather forecasting models have enhanced TC track prediction, they often underestimate TC intensity \cite{fuxi_extreme2024,xu2025exploring}.
Operational forecasting centers typically address such biases through human forecaster adjustments or objective post‑processing, combining qualitative, case-specific edits with quantitative corrections.
This process is analogous to image editing \cite{Zhan2023,fu2024}, in which weather forecasts are treated as images whose key attributes are modified while preserving physical consistency.
Similarly, FuXi-Uni aligns textual and Earth science data to “edit” TC forecasts by enhancing intensity while preserving physically consistent relationships among wind, pressure and other atmospheric fields.
This capability demonstrates the potential of FuXi-Uni as an interface that integrates human forecaster expertise with quantitative post‑processing, translating natural language instructions into physically consistent, bias-corrected forecast adjustments extending beyond TC intensity alone.

\begin{figure}[h]
\centering
\includegraphics[width=\linewidth]{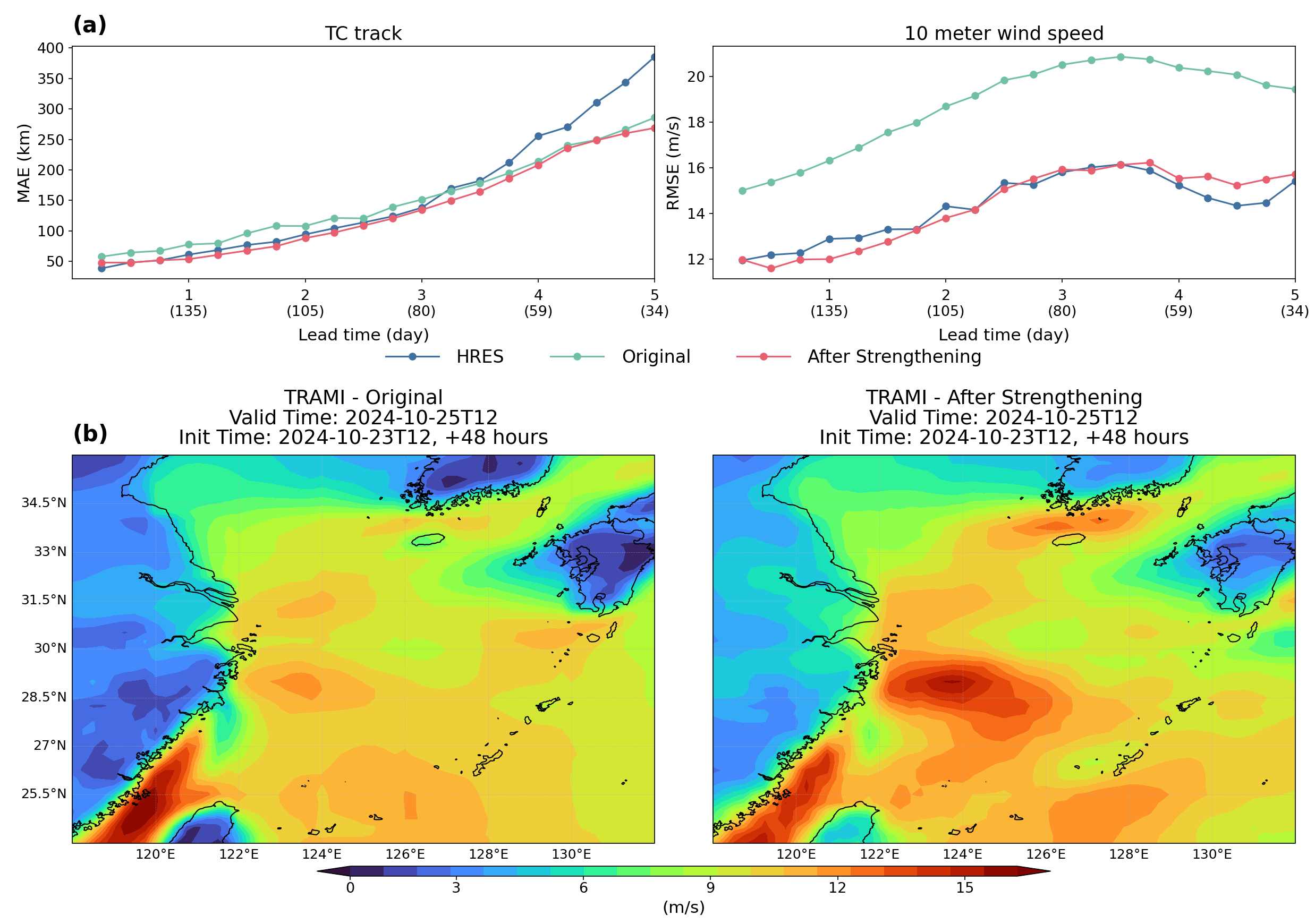}
\caption{\textbf{FuXi-Uni improves both tropical cyclone (TC) track and intensity forecasts.} \textbf{a,} Mean absolute error (MAE) of TC track forecasts (first column) and root mean square error (RMSE) of 10-meter wind speed ($\textrm{WS10M}$, second column) for TC intensity forecasts as a function of forecast lead times, comparing ECMWF HRES (blue lines), original FuXi-Uni (green lines), and FuXi-Uni with strengthened intensity (red lines). Evaluation is based on forecasts of 20 TCs and uses IBTrACS as the reference. \textbf{b,}  Comparison of predicted TC intensity and structure for Typhoon Trami (2420). The figure shows the spatial distribution of $\textrm{WS10M}$ (m/s) at 12 UTC on 25 October 2024 from original FuXi-Uni (first column) and FuXi-Uni with strengthened intensity (second column), initialized at 12 UTC on 23 October 2024.}
\label{TC_edit}
\end{figure}

Figure \ref{TC_edit}a presents time series of TC track MAE and intensity error, quantified by WS10M RMSE, for forecasts from ECMWF HRES, the original FuXi-Uni, and the intensity-strengthened FuXi-Uni.
The evaluation includes 20 TCs (see Table \ref{summary_TC}) that occurred between May and October 2024.
Prior to intensity adjustment, FuXi-Uni, consistent with other AI models, shows superior track forecasts relative to ECMWF HRES, with the advantage increasing with lead time, but poorer intensity forecasts, reflected by by larger WS10M RMSE across the 5-day forecasts.
After intensity strengthening, FuXi-Uni exhibits a clear improvement in intensity prediction, achieving a slightly lower overall WS10M RMSE than ECMWF HRES (13.840 m/s vs 13.930 m/s).
The strengthened FuXi-Uni also modestly improves track forecasts, reducing the track MAE from 123.3 km (original FuXi-Uni) to 106.2 km.

Beyond statistical metrics, Figure \ref{TC_edit}b examines the spatial structure of the strengthened TC, showing the spatial map of $\textrm{WS10M}$ (m/s) for Typhoon Trami (2420) at 12 UTC on 25 October 2024 from the original FuXi-Uni and strengthened FuXi-Uni forecasts, with warmer colors indicating higher wind speeds.
In late October 2024, Typhoon Trami (known in the Philippines as Kristine), struck the northern Philippines, producing persistent torrential rainfall and catastrophic flooding that caused over overall 160 fatalities and affected millions of people \cite{Qian2025}.
Its slow movement and orographic enhancement led to a maximum cumulative rainfall of 1243.1 mm in Qionghai, Hainan, China, triggering landslides and extensive damage to housing, infrastructure and agriculture, with economic losses reaching hundreds of millions of U.S. dollars \cite{Trami_HKO,merz2024climate}.
As shown in Figure \ref{TC_edit}b, the TC intensity enhancement is not limited to the TC center as wind speeds are increased across the entire storm circulation, consistent with the large spatial scale of TCs, which typically span several hundred kilometers in diameter.
Note that because FuXi-Uni does not always underestimate TC intensity, the results shown are based on filtered testing dataset that includes only cases where FuXi-Uni underestimates intensity relative to IBTrACS and the target dataset (see \ref{WRF_dataset}).


\FloatBarrier

\subsubsection{Spatial downscaling}

This subsection shows an additional example of FuXi-Uni as a natural language-driven interface for performing spatial downscaling, extending beyond tropical cyclone intensity adjustment.
The task addresses a common practical need, as many applications require high-resolution information derived from coarse-resolution model outputs.

When prompted to downscale the regional weather state from 1.5$^\circ$ to 0.25$^\circ$ resolution, FuXi-Uni produces high-resolution fields within seconds on a single GPU.
In contrast, conventional regional NWP models, such as the Weather Research and Forecasting (WRF) model \cite{jensen2021description}, typically require several hours of calculation on hundreds to thousands of CPU cores.

\begin{figure}[h]
\centering
\includegraphics[width=\linewidth]{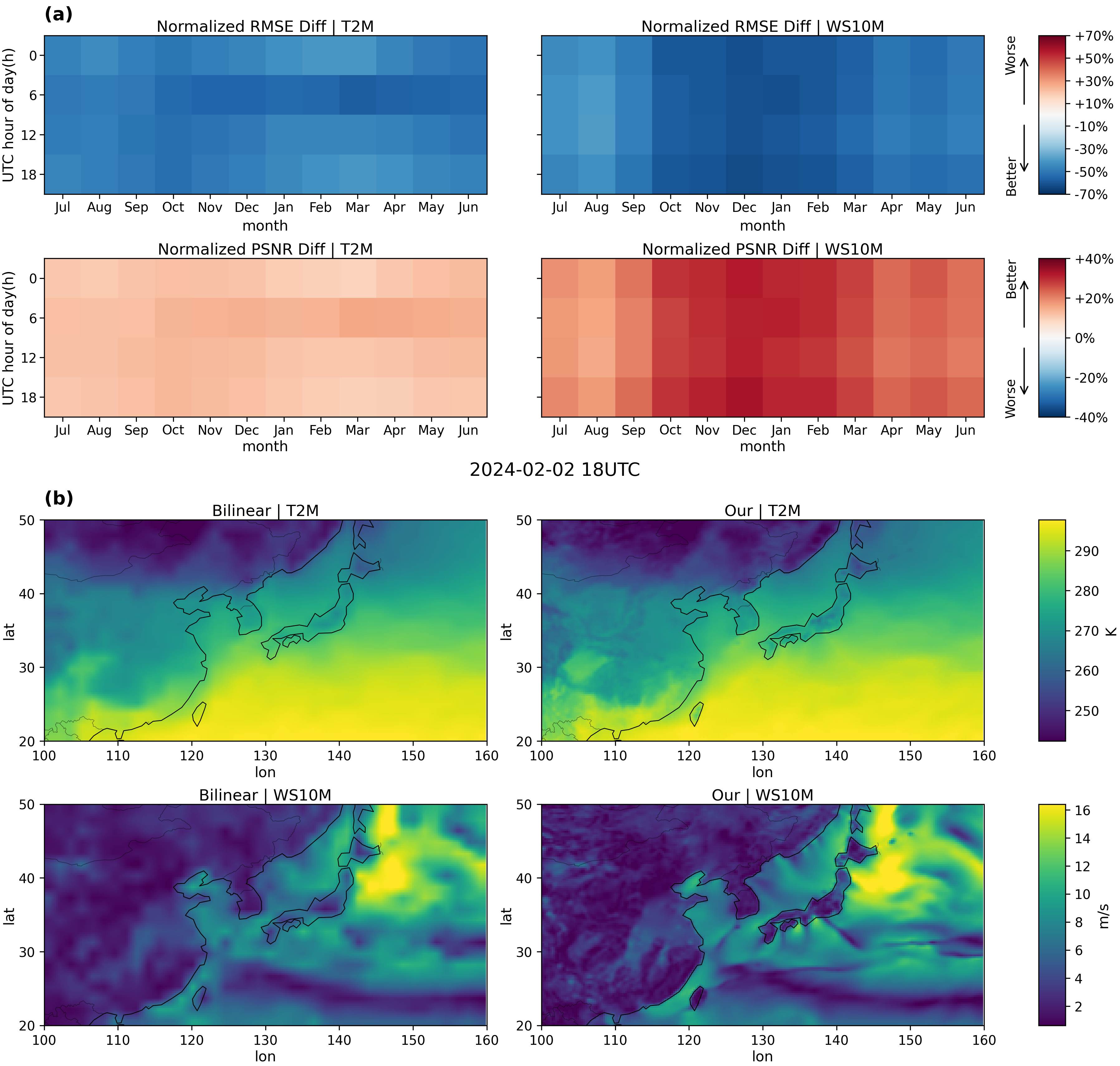}
\caption{\textbf{FuXi-Uni outperforms bilinear interpolation in terms of downscaling ERA5 reanalysis from 1.5$^\circ$ to 0.25$^\circ$.} \textbf{a,} Comparison of normalized differences in root mean squared error ($\textrm{RMSE}$, first row) and peak signal-to-noise ratio (PSNR, second row) of FuXi-Uni compared to bilinear interpolation in 0.25$^\circ$ resolution fields downscaled from 1.5$^\circ$ resolution ERA5. The x-axis and y-axis correspond to month of the year and UTC hour of day. Results are shown for two variables: 2-meter temperature ($\textrm{T2M}$, first row) and and 10-meter wind speed ($\textrm{WS10M}$, second row), calculated using all testing data over a 1-year testing period (June 01, 2023 - June 30, 2024). \textbf{a,} Comparison of example downscaled results from bilinear interpolation and FuXi-Uni at 18 UTC on February 02, 2024 for ($\textrm{T2M}$. first row) and ($\textrm{WS10M}$, second row).}
\label{spatial_downscaling}
\end{figure}

Figure \ref{spatial_downscaling}a illustrates the normalized differences in $\textrm{RMSE}$ and peak signal-to-noise ratio (PSNR) for T2M and WS10M downscaled by FuXi-Uni and bilinear interpolation, using 0.25$^\circ$ ERA5 as the reference.
Both methods take 1.5$^\circ$ ERA5 as input, with bilinear interpolation serving as the baseline for the normalized difference calculation (see Section \ref{evaluation_method}).
The evaluation covers a 1-year testing period from June 01, 2023 to June 30, 2024.
For RMSE, blue, red, and white denote regions where FuXi-Uni achieves lower, higher, or comparable errors relative to bilinear interpolation, respectively.
For PSNR, the same color scheme denotes lower, higher, or comparable reconstruction fidelity.
FuXi-Uni consistently outperforms bilinear interpolation across all hours and months for both metrics.
Figure \ref{spatial_downscaling}b presents an example snapshot of downscaled T2M and WS10M produced by FuXi-Uni and bilinear interpolation, respectively at 18 UTC on February 2, 2024.
FuXi-Uni reproduces substantially richer fine-scale spatial structures than bilinear interpolation.

\FloatBarrier

\subsection{Biomedical visual question answering}


Biomedical VQA is a multimodal AI task in which a model interprets a natural language clinical question about a medical image and generates an accurate, clinically relevant answer \cite{lin2021medical,HU2024}.
Its significance lies in the potential to democratize medical expertise by providing a scalable, tireless assistant for clinical decision support, medical education, and patient interaction.
Progress in biomedical VQA has been driven by advances from simple neural networks to modern multimodal transformers and LMMs, together with the development of domain-specific benchmarks.
Early datasets such as VQA-RAD \cite{VQA_RAD} laid the foundation by providing high-quality, clinician-validated question-answer pairs for radiology images.
Subsequent efforts, including PathVQA \cite{PathVQA}, expanded the coverage to pathology through large-scale, semi-automated generation based on medical image captions, and SLAKE \cite{SLAKE}, which enriched the field with bilingual annotations and semantic labels (e.g., segmentation masks) to enhance visual grounding.

Methodological advances have been largely driven by LMMs and domain adaptation strategies.
For instance, LLaVA-Med \cite{li2023llava} adapts a general vision-language assistant to the biomedical domain through multimodal instruction tuning on medical images and texts, leading to improved performance on standard biomedical VQA benchmarks.
To further improve robustness and generalization, MedTrinity-25M \cite{xie2024medtrinity} scales up multimodal medical pretraining using multigranular annotations spanning 10 modalities and over 65 diseases, yielding further gains in downstream biomedical VQA tasks and enabling better models such as LLaVA-Tri.
This subsection evaluates the performance of FuXi-Uni on biomedical VQA, which is distinct from the Earth science tasks discussed in previous sections.
Details of the evaluation metrics are provided in Section~\ref{evaluation_VQA}.

Table \ref{biomed_vqa_results} presents results on VQA-RAD, SLAKE, and PathVQA.
We include LLaVA \cite{liu2023visual} as a general-domain baseline, along with two recent strong biomedical VQA models, LLaVA-Med and LLaVA-Tri.
Both are built upon the LLaVA backbone and enhanced via medical domain adaptation or pretraining, with LLaVA-Tri further leveraging large-scale multigranular medical data from MedTrinity \cite{li2023llava,xie2024medtrinity}.
In contrast, FuXi-Uni is built on Qwen2.5-VL \cite{bai2025qwen2}, a stronger general-purpose vision-language foundation model than the original LLaVA backbone.
While part of FuXi-Uni’s performance gains can be attributed to this improved foundation model.

Compared with the LLaVA-based biomedical SOTA models, FuXi-Uni consistently achieves superior performance across all evaluated benchmarks.
Relative to LLaVA-Med, FuXi-Uni outperform it on all six metrics, with particularly notable improvements on VQA-RAD and SLAKE (e.g., +6.7/+3.1 on VQA-RAD Open/Closed and +4.1/+4.1 on SLAKE Open/Closed).
Against LLaVA-Tri, FuXi-Uni achieves the better overall performance on VQA-RAD and PathVQA, while remaining competitive on SLAKE and slightly improving both Open and Closed accuracy.
These results indicate that FuXi-Uni not only benefits from a stronger backbone, but also delivers additional domain-specific gains beyond existing LLaVA-based medical adaptation strategies, improving both open-ended clinical reasoning and closed-ended diagnostic accuracy.
Nevertheless, a substantial performance gap remains between open-ended and closed-ended questions, suggesting that biomedical VQA, especially for open-ended reasoning, remains challenging and require larger or more diverse training data \cite{VQA_RAD}.

\newlength{\qaimgH}
\setlength{\qaimgH}{1.8cm}                 

\newlength{\qacapH}
\setlength{\qacapH}{4.0\baselineskip}      

\newlength{\qaW}
\setlength{\qaW}{0.18\textwidth}           

\newcommand{\qaimg}[4][\qaimgH]{%
  \begin{minipage}[t]{\qaW}\vspace{0pt}
    \centering
    \parbox[t][#1][c]{\linewidth}{%
      \
      \adjustbox{valign=t}{%
          \includegraphics[width=\linewidth,height=\qaimgH,keepaspectratio]{\detokenize{#2}}%
        }
    }\\[8pt] 
    \parbox[t][\qacapH][t]{\linewidth}{%
      \centering\small
      \textbf{Question:}\\[-1pt]
      \textbf{#3}~#4%
    }
  \end{minipage}%
}

\newcolumntype{R}[1]{>{\raggedright\arraybackslash\small}p{#1}}
\newcolumntype{C}[1]{>{\centering\arraybackslash\small}p{#1}}
\newcolumntype{A}{>{\raggedright\arraybackslash\small\hsize=0.95\hsize}X} 
\newcolumntype{B}{>{\raggedright\arraybackslash\small\hsize=1.75\hsize}X} 
\newcolumntype{D}{>{\raggedright\arraybackslash\small\hsize=1.45\hsize}X} 

\begin{figure*}[t]
\centering

\qaimg{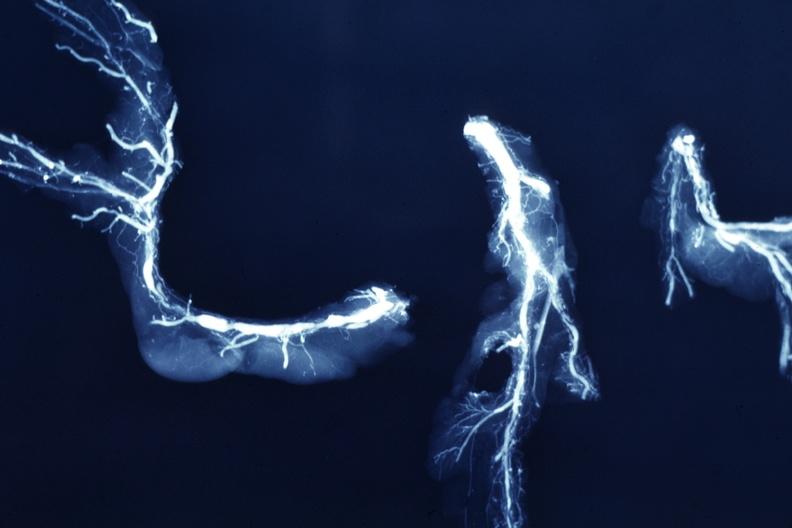}{(1)}{what is present?}\hfill
\qaimg{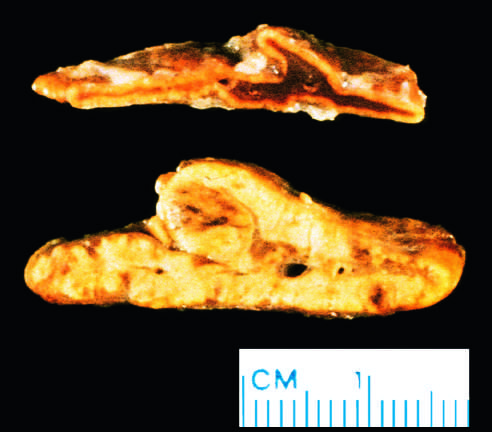}{(2)}{what was the abnormal gland from?}\hfill
\qaimg{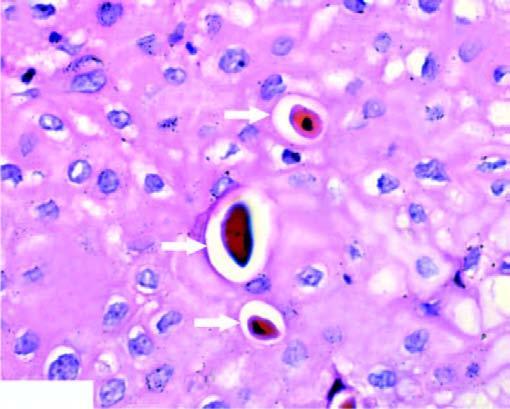}{(3)}{Is the dead cell seen in singles?}\hfill
\qaimg{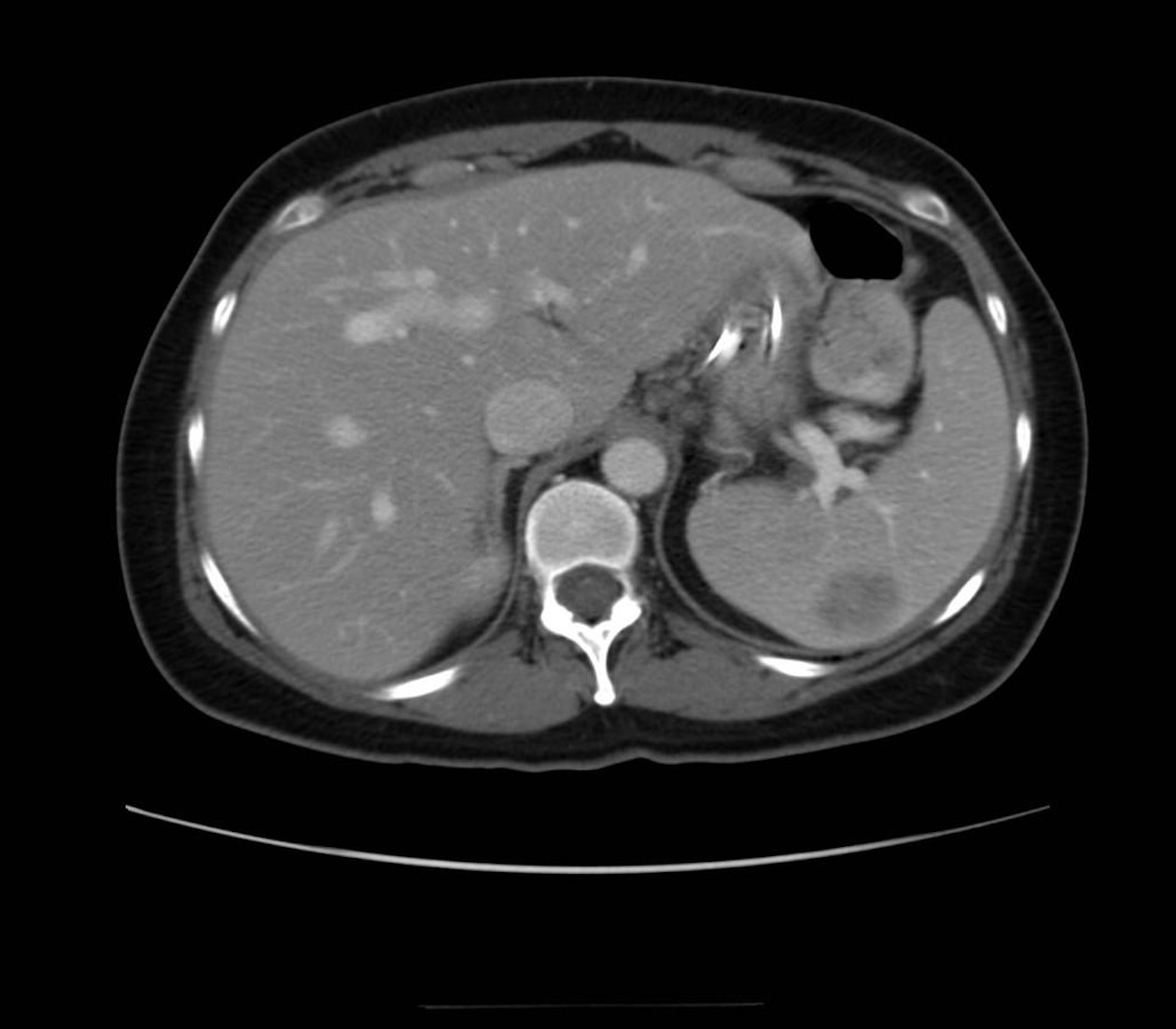}{(4)}{How would you describe the spleen abnormality?}\hfill
\qaimg{}{(5)}{What side of the brain is a lesion on?}

\vspace{14pt} 

\renewcommand{\arraystretch}{1.20}
\setlength{\tabcolsep}{5pt} 

\begin{tabularx}{\textwidth}{@{}%
  >{\raggedright\arraybackslash\small\bfseries}p{0.15\textwidth} 
  >{\raggedright\arraybackslash\small}X                         
  >{\raggedright\arraybackslash\small}X                         
  >{\centering\arraybackslash\small}p{0.07\textwidth}            
  >{\raggedright\arraybackslash\small}X                         
  >{\centering\arraybackslash\small}p{0.07\textwidth}            
@{}}
\toprule
 & \multicolumn{1}{c}{(1)} & \multicolumn{1}{c}{(2)} & \multicolumn{1}{c}{(3)} &
   \multicolumn{1}{c}{(4)} & \multicolumn{1}{c}{(5)} \\
\midrule

Answer &
vasculature &
\makecell[l]{a patient with\\ ACTH\mbox{-}dependent\\ Cushing syndrome} &
yes &
\makecell[l]{right lower lateral\\ lung field} &
Left \\

\textcolor{blue}{LLaVA\_Tri} &
\makecell[l]{coronary\\ artery} &
endocrine system &
no &
nothing &
Right \\

\textcolor{red}{FuXi-Uni} &
vasculature &
\makecell[l]{ACTH\mbox{-}dependent\\ Cushing} &
yes &
hypodense lesion &
Left \\
\bottomrule
\end{tabularx}

\caption{Representative cases sampled from VQA-RAD and PathVQA. For each image--question pair (1--5), we list the ground-truth answer and the corresponding predictions produced by LLaVA\_Tri and FuXi-Uni.}

\label{fig:vqa_rad_example}
\end{figure*}

To provide an intuitive comparison, we present qualitative examples sampled from both VQA-RAD and PathVQA in Figure \ref{fig:vqa_rad_example}. Each column shows an image--question pair (top) and the corresponding ground-truth answer together with model predictions (bottom). Overall, the compared methods exhibit noticeable differences across question types such as anatomical structure identification, attribute description, and laterality reasoning. In these representative cases,FuXi-Uni more frequently matches the ground truth, whereas LLaVA-Tri occasionally produces concept confusions or incorrect laterality judgments.

\begin{table}[h]
\centering
\caption{\textbf{FuXi-Uni outperforms leading models in terms of biomedical visual question answering (VQA) accuracy.} Biomedical VQA performance of GPT-4V, LLaVA, LLaVA-Med, LLaVA-Tri, Qwen2.5-VL, and FuXi-Uni on three benchmarks: VQA-RAD, SLAKE, and PathVQA. Results are presented separately for open-ended and closed-ended questions following the standard protocol of each benchmark. Bold font suggests the highest accuracy.}
\label{biomed_vqa_results}
\begin{tabular}{l|cc|cc|cc}
\toprule
\multirow{2}{*}{Method} &
\multicolumn{2}{c|}{VQA-RAD} &
\multicolumn{2}{c|}{SLAKE} &
\multicolumn{2}{c}{PathVQA} \\
& Open & Closed & Open & Closed & Open & Closed \\
\midrule
GPT-4V \cite{openai2024gpt4}          & 39.5 & 78.9 & 33.6 & 43.6 & --  & --  \\
LLaVA \cite{liu2023visual}           & 50.0 & 65.1 & 78.2 & 63.2 & 7.7 & 63.2 \\
LLaVA-Med \cite{li2023llava}         & 61.5 & 84.2 & 83.1 & 85.3 & 37.9 & 91.2 \\
LLaVA-Tri \cite{xie2024medtrinity}  & 65.1 & 84.9 & 86.4 & 89.2 & 37.9 & 92.7 \\
Qwen2.5-VL \cite{bai2025qwen2}       & 53.2 & 79.7 & 80.4 & 64.1 & 8.5  & 64.9  \\
FuXi-Uni                              & \textbf{68.18} & \textbf{87.32} & \textbf{87.17} & \textbf{89.4} & \textbf{39.1} & \textbf{93.05} \\
\bottomrule
\end{tabular}
\end{table}



\section{Discussion}

The rapid growth and increasing heterogeneity of scientific data, together with global interdisciplinary challenges, call for unified multimodal AI models capable of working seamlessly across scientific domains.
However, most existing models either remain limited to single domains or rely on text-centric tokenization, which is ill-suited to high-dimensional scientific data.
Here, we introduce FuXi-Uni, a multimodal AI model that integrates high-dimensional data from multiple scientific domains within a single architecture, enabling both understanding and generation through natural language instructions.
By aligning domain-specific scientific tokens with textual representations in a shared latent space, FuXi-Uni preserves the structural integrity of complex data, including spatiotemporal weather fields and biomedical images.
Across Earth science and biomedicine, FuXi-Uni matches or surpasses SOTA domain-specific physical and AI models.

In Earth system modeling, FuXi-Uni generates 10-day global weather forecasts at 0.25$^\circ$ spatial and 6-hour temporal resolution, outperforming ECMWF HRES, the world's leading operational NWP system.
Prompt-based post-processing enhances underestimated TC intensities while maintaining physical consistency, making FuXi-Uni the first AI model to surpass ECMWF HRES in both TC track and intensity prediction.
For spatial downscaling from 1.5$^\circ$ to 0.25$^\circ$, FuXi-Uni outperforms bilinear interpolation while resolving finer-scale structures.
Together, these capabilities unify forecasting, bias correction and downscaling within a single framework, streamlining operational workflows and substantially reducing computational costs relative to physical models.
Thus, FuXi-Uni functions as an "AI meteorological forecaster" that, for the first time, integrates human forecaster expertise with quantitative post-processing in a unified model.
In biomedicine, FuXi-Uni achieves SOTA performance in VQA across the VQA-RAD, SLAKE and PathVQA benchmarks, surpassing leading multimodal AI models such as LLaVA-Med and LLaVA-Tri.
Instruction tuning on merged datasets enables robust multimodal generalization without catastrophic forgetting, highlighting the scalability of the unified architecture across biomedical tasks.

Furthermore, FuXi-Uni establishes a new paradigm  for general-purpose scientific AI model.
It demonstrates that a unified, science-aware multimodal architecture can simultaneously advance Earth system prediction and biomedical image understanding while maintaining a natural language interface.
Rather than collapsing all modalities into text, FuXi-Uni aligns domain-specific scientific tokens with textual representations in a shared latent space, supporting diverse scientific tasks through natural language instructions.
Future extensions may incorporate additional domains and tasks, where joint multi-domain training could further enhance cross-domain transfer.
By unifying scientific modalities under a language-based interface, FuXi-Uni has the potential to empower researchers to tackle grand interdisciplinary challenges, such as climate tipping points, pandemics, and sustainable energy transitions, promoting collaborative, system-level advances to solving those global challenges.

If the 2020–2024 AlphaFold revolution marked the moment when AI first helped human read the book of life, the emergence of unified multimodal understanding and generation models represents a quieter yet deeper shift, one in which AI begins to accompany us as we revisit, annotate and cautiously extend its pages.
The era of multimodal foundation models in science has begun, but as a additional way in how we ask questions, test hypotheses and imagine what may become possible.

\section{Methods}

\subsection{Dataset for Earth science}

\subsubsection{ERA5 reanalysis dataset}

ERA5 is the fifth generation atmospheric reanalysis dataset produced by ECMWF.
It provides hourly data of the global atmosphere, land surface, and ocean waves from January 1940 to the present, with a horizontal resolution of approximately 31 km.
ERA5 is generated using a four-dimensional variational data assimilation system within Cycle 41r2 of ECMWF’s Integrated Forecasting System (IFS), which was operational for most of time in 2016 \cite{hersbach2020era5}.

FuXi-Uni includes 70 meteorological variables, including 5 upper-air atmospheric variables across 13 pressure levels (50, 100, 150, 200, 250, 300, 400, 500, 600, 700, 850, 925, and 1000 hPa), and 5 surface variables.
Upper-air variables include geopotential (${\textrm{Z}}$), temperature (${\textrm{T}}$), u component of wind (${\textrm{U}}$), v component of wind (${\textrm{V}}$), and specific humidity (${\textrm{Q}}$).
Surface variables are 2-meter temperature (${\textrm{T2M}}$), mean sea-level pressure (${\textrm{MSL}}$), 10-meter u wind component (${\textrm{U10M}}$), 10-meter v wind component (${\textrm{V10M}}$), and 10-meter wind speed (${\textrm{WS10M}}$).
A comprehensive list of these variables and their abbreviations is detailed in Table \ref{glossary_weather}. 

\begin{table}
\centering
\caption{\label{glossary_weather} A summary of input and output variables for global weather forecasting and spatial downscaling. The "Type" column whether a variable is a time-varying including upper-air, surface, and geographical variables, or temporal. The “Full name” and “Abbreviation” columns list each variable's complete name and its abbreviations as used in this paper. The "Role" column specifies whether a variable is used as both input and output, or as input only. "I \& O" denotes variables used as input and output, and "I" suggests variables used as input only.}
\begin{tabularx}{\textwidth}{cXccc}
\hline
\textbf{Type} & \textbf{Full name} & \textbf{Abbreviation} & \textbf{Role} \\
\hline
upper-air           & geopotential & ${\textrm{Z}}$ & I \& O \\
                    & temperature  & ${\textrm{T}}$ & I \& O \\
                    & u component of wind & ${\textrm{U}}$ & I \& O \\
                    & v component of wind & ${\textrm{V}}$ & I \& O \\
                    & specific humidity & ${\textrm{Q}}$ & I \& O \\
                    \hline
surface             & 2-meter temperature & ${\textrm{T2M}}$ & I \& O \\
                    & mean sea-level pressure & ${\textrm{MSL}}$ & I \& O \\
                    & 10-meter u wind component & ${\textrm{U10M}}$ & I \& O \\
                    & 10-meter v wind component & ${\textrm{V10M}}$ & I \& O \\
                    & 10-meter wind speed & ${\textrm{WS10M}}$ & I \& O \\ 
\hline
geographical        & orography & ${\textrm{OR}}$ & I \\
                    & land-sea mask & ${\textrm{LSM}}$ & I \\
                    & latitude & ${\textrm{LAT}}$ & I \\
                    & longitude & ${\textrm{LON}}$ & I \\  
\hline                  
temporal            & hour of day & ${\textrm{HOUR}}$ & I \\
                    & day of year & ${\textrm{DOY}}$ & I \\
                    & step & ${\textrm{STEP}}$ & I \\                   
\hline
\end{tabularx}
\end{table}

\subsubsection{WRF dataset}
\label{WRF_dataset}

We use the Weather Research and Forecasting (WRF) dataset produced by Guo et al. \cite{guo2025fuxitcg} as the basis for improving TC forecasts of FuXi-Uni.

\begin{figure}[h]
    \centering
    \includegraphics[width=\linewidth]{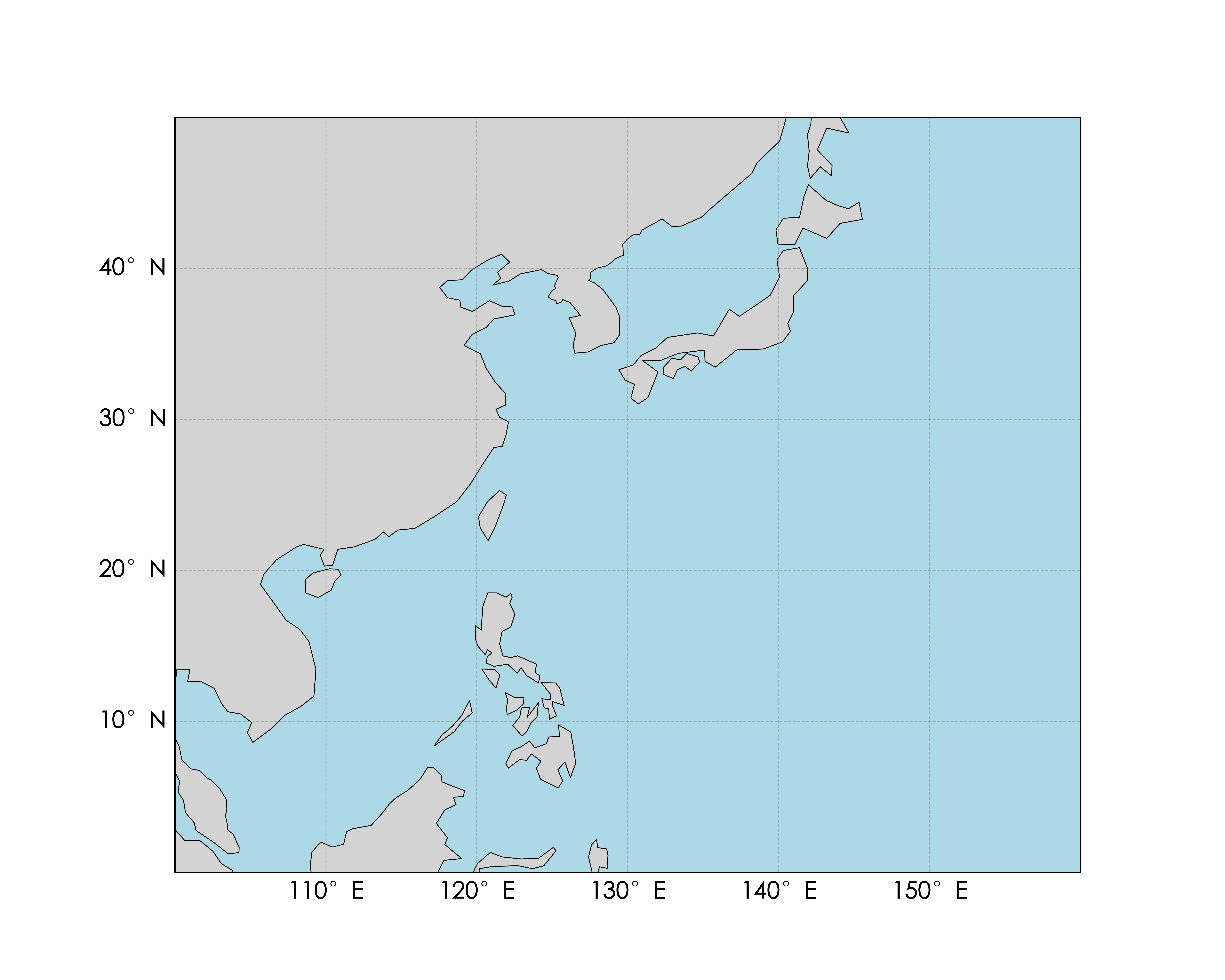}
    \caption{WRF simulation domain over the western North Pacific at 0.25$^\circ$ spatial resolution.}
    \label{wrf_domain}
\end{figure}

Regional simulations are performed with WRF version 4.3 \cite{jensen2021description} over the western North Pacific (WNP).
The model is initialized twice daily at 00 and 12 UTC, with forecast lead times extending up to 120 hours.
A latitude–longitude projection is used to support numerical integration and subsequent AI-based post-processing.
Model outputs are generated at 0.25$^\circ$ spatial resolution for the period 2019–2024, with simulations from 2019–2023 used for model training and those from 2024 reserved exclusively for evaluation.
The simulation domain covers 100$^\circ$E–160$^\circ$E and 0$^\circ$N–50$^\circ$N (Figure \ref{wrf_domain}), corresponding to a 242 $\times$ 202 grid at 0.25$^\circ$ resolution.
The model top is set at 50 hPa and has 56 model levels.

\FloatBarrier

In addition, upper-air fields are interpolated to standard pressure levels using linear interpolation.
From the WRF forecasts, we extract a set of TC-relevant predictors, including five upper-air atmospheric variables at five pressure levels (200, 300, 500, 700 and 850 hPa), together with surface meteorological variables listed in Table \ref{glossary_weather}.
These fields serve as inputs to the AI model, enabling improved TC intensity prediction and more accurate representation of vertical storm structure.

\begin{table}[h]
\centering
\caption{\label{summary_TC} List of the 20 tropical cyclones (TCs) evaluated in this study, including their names and the forecast initialization start and end times (UTC).}
\begin{tabular}{lll}
\toprule
TC\_name & init\_time & end\_time \\
\midrule
EWINIAR   & 2024-05-25 00:00:00 & 2024-05-30 12:00:00 \\
MALIKSI   & 2024-05-31 00:00:00 & 2024-05-31 12:00:00 \\
PRAPIROON & 2024-07-20 00:00:00 & 2024-07-22 12:00:00 \\
GAEMI     & 2024-07-20 00:00:00 & 2024-07-25 00:00:00 \\
MARIA     & 2024-08-07 00:00:00 & 2024-08-12 12:00:00 \\
AMPIL     & 2024-08-12 12:00:00 & 2024-08-18 00:00:00 \\
SON-TINH  & 2024-08-12 12:00:00 & 2024-08-13 00:00:00 \\
WUKONG    & 2024-08-13 00:00:00 & 2024-08-14 12:00:00 \\
JONGDARI  & 2024-08-19 00:00:00 & 2024-08-20 00:00:00 \\
SHANSHAN  & 2024-08-21 12:00:00 & 2024-09-01 00:00:00 \\
YAGI      & 2024-09-01 12:00:00 & 2024-09-07 00:00:00 \\
LEEPI     & 2024-09-04 00:00:00 & 2024-09-06 12:00:00 \\
BEBINCA   & 2024-09-10 00:00:00 & 2024-09-15 12:00:00 \\
PULASAN   & 2024-09-17 00:00:00 & 2024-09-21 00:00:00 \\
SOULIK    & 2024-09-18 12:00:00 & 2024-09-19 00:00:00 \\
CIMARON   & 2024-09-24 12:00:00 & 2024-09-27 00:00:00 \\
JEBI      & 2024-09-26 12:00:00 & 2024-10-01 12:00:00 \\
KRATHON   & 2024-09-27 12:00:00 & 2024-10-03 00:00:00 \\
BARIJAT   & 2024-10-06 12:00:00 & 2024-10-10 00:00:00 \\
TRAMI     & 2024-10-20 12:00:00 & 2024-10-28 12:00:00 \\
KONG-REY  & 2024-10-25 00:00:00 & 2024-10-31 12:00:00 \\
\bottomrule
\end{tabular}
\end{table}

\subsubsection{WRF dataset filtering}
\label{WRF_filter}

To pair FuXi-Uni with WRF data, we subset cases in which WRF predicts stronger tropical cyclones (TCs) than FuXi-Uni.
The data filtering is based on the mean bias error (MBE) of predicted maximum 10-m wind speed (WS10M), using IBTrACS data as the reference.
WRF forecasts are excluded as training targets when the FuXi-Uni MBE is smaller than that of WRF.
When the FuXi-Uni MBE exceeds that of WRF and both models underestimate WS10M relative to IBTrACS, TC intensity is enhanced after TC editing.
On the contrary, when both models overestimate WS10M relative to IBTrACS, the intensity is weakened.
As an additional constraint, cases in which FuXi-Uni and WRF exhibit comparable WS10M MBEs but the track position error exceeds a prescribed threshold (10 km in this study) are also excluded.


\subsubsection{Tropical cyclone dataset}

We conducted assessments of TC forecasts using the IBTrACS \cite{knapp2010,gahtan2024international} dataset as the reference, which is provided by the National Oceanic and Atmospheric Administration (NOAA). IBTrACS combines all accessible best track datasets from around the world into a comprehensive compilation. Each track in the dataset represents a 6-hourly time series of a TC's eye location in terms of latitude and longitude coordinates, along with other relevant features at that specific time and location. In alignment with established practices for evaluating TC predictions \cite{Magnusson2021}, we evaluate all TC tracks when original FuXi-Uni, FuXi-, and HRES concurrently detect a cyclone. This approach ensures that all models are evaluated using the same set of events.

To facilitate a comparison with ECMWF HRES, we used the THORPEX Interactive Grand Global Ensemble (TIGGE) \cite{bougeault2010,swinbank2016tigge} archive, which contains cyclone tracks estimated using the operational ECMWF tracker.
The ECMWF TC track data, stored in XML file format, include TC tracks derived from both ECMWF HRES and ensemble forecasts. We specifically extract the HRES forecasts based the "forecast" tag.

In addition to the IBTrACS dataset, we implemented the Aurora TC tracker \cite{bodnar2025foundation} to the ERA5 dataset to extract TC tracks and intensity for TC forecast evaluations.

\subsection{Dataset for biomedicine}

\subsubsection{Biomedical Pretraining Dataset}

In our experiments, we adopt the MedTrinity-25M dataset proposed by Xie et al \cite{xie2024medtrinity}., which is currently one of the largest and most richly annotated open-source multimodal medical datasets. MedTrinity-25M contains about 25 million {image, ROI, description} triplets collected from more than 90 public medical datasets and online repositories. The images cover 10 imaging modalities, including MRI, CT, X-ray, histopathology, endoscopy, ultrasound, PET, dermoscopy, and microscopy, and span a wide range of anatomical regions (brain, thorax, abdomen, pelvis, etc.) and over 65 diseases. A key feature of MedTrinity-25M is its multigranular annotation scheme: for each image, the dataset provides one or more regions of interest (ROIs) in the form of bounding boxes or segmentation masks, together with a structured textual description. The description combines global information (imaging modality, primary organ, disease/lesion type) with fine-grained local information (ROI location, area ratio, signal or intensity changes, texture patterns) and explicitly models the relationship between local abnormalities and the global organ status. Compared with traditional medical datasets that only provide image–report pairs or coarse classification labels, MedTrinity-25M offers much richer supervision in both spatial and textual dimensions. Pretraining medical vision–language models on MedTrinity-25M has been shown to substantially improve downstream performance on benchmarks such as VQA-RAD \cite{VQA_RAD}, SLAKE \cite{SLAKE}, and PathVQA \cite{PathVQA}, making it a suitable large-scale data source for both pretraining and task-specific finetuning. 

\subsubsection{Biomedical finetuning and evaluation datasets}

For downstream biomedical evaluation, we follow Xie et al.\ and fine-tune and assess our model on three standard Med-VQA benchmarks: VQA-RAD, SLAKE, and PathVQA.

\begin{itemize}
\setlength\itemsep{1em}
    \item \textbf{VQA-RAD}:
    VQA-RAD is a radiology VQA dataset constructed from MedPix, containing 315 de-identified CT/MRI/X-ray images and 3,515 clinician-authored question–answer (QA) pairs; questions cover imaging modality, anatomical region, and the presence and type of abnormality, with answers mixing binary yes/no and short open-ended text. The official protocol designates 451 QA pairs as a held-out test set, with the remaining samples used for training (and, in our case, a small validation split).
    
    \item \textbf{SLAKE}:
    SLAKE is a semantically labeled, knowledge-enhanced bilingual Med-VQA dataset with 642 CT/MRI/X-ray images, dense visual annotations, an associated medical knowledge base, and 14,028 QA pairs in Chinese and English; images are split at the image level into 70\%/15\%/15\% train/validation/test, and questions are explicitly categorized into purely visual and knowledge-based types, enabling separate assessment of perceptual understanding and medical reasoning.
    
    \item \textbf{PathVQA}:
    PathVQA targets pathology images and consists of 4,998 textbook- and PEIR-derived histopathology images and 32,799 QA pairs, partitioned into training, validation, and test sets with a ratio of roughly 3:1:1 (19,755 / 6,279 / 6,761 QA); questions include both yes/no and free-form open-ended items that resemble board-style pathology exam questions, covering structure recognition, lesion morphology, counting, and diagnostic judgments. Across all three benchmarks, we adopt a unified Med-VQA setting in which the model receives an image–question pair and generates a short textual answer; evaluation is performed on the standard train/validation/test splits using exact-match top-1 accuracy as the main metric, and we additionally report performance on open-ended vs.\ closed-ended subsets where applicable.

\end{itemize}

\subsection{Evaluation method}
\label{evaluation_method}

\subsubsection{Evaluation method for Earth science}

All weather forecasts are evaluated against benchmark datasets at corresponding valid times.
For original or revised FuXi-Uni forecasts initialized from ERA5, ERA5 reanalysis data is used as the reference.
For ECMWF high-resolution (HRES) forecasts, the operational analysis time series (HRES-fc0) used for model initialization at time $t_0$ serves as the reference at the corresponding valid time $t_0+\tau$.
 
Deterministic forecast skill is assessed using standard metrics, including the globally-averaged and latitude-weighted root mean square error (RMSE) and anomaly correlation coefficient (ACC), defined as:

\begin{equation}
\label{RMSE_equation}
    \textrm{RMSE}(c, \tau) =\frac{1}{\mid \textrm{D} \mid}\sum_{t_0 \in \textrm{D}} \sqrt{\frac{1}{\textrm{H} \times \textrm{W}} \sum_{i=1}^\textrm{H}\sum_{j=1}^{\textrm{W}} a_i {( \hat{\textrm{\textbf{X}}}^{t_0 +\tau}_{c,i,j} - \textrm{\textbf{X}}^{t_0 +\tau}_{c,i,j} )}^{2}}
\end{equation}

\begin{equation}
\label{ACC_equation}
    \textrm{ACC}(c, \tau) = \frac{1}{\mid \textrm{D} \mid}\sum_{t_0 \in \textrm{D}} \frac{\sum_{i, j} a_i (\hat{\textrm{\textbf{X}}}^{t_0 +\tau}_{c,i,j} - \textrm{\textbf{M}}^{t_0 +\tau}_{c,i,j}) (\textrm{\textbf{X}}^{t_0 +\tau}_{c,i,j} - \textrm{\textbf{M}}^{t_0 +\tau}_{c,i,j})} {\sqrt{ \sum_{i, j} a_i (\hat{\textrm{\textbf{X}}}^{t_0 +\tau}_{c,i,j} - \textrm{\textbf{M}}^{t_0 +\tau}_{c,i,j})^2 \sum_{i, j} a_i(\textrm{\textbf{X}}^{t_0 +\tau}_{c,i,j} - \textrm{\textbf{M}}^{t_0 +\tau}_{c,i,j})^2}}
\end{equation}

where, ${t_0}$ denotes the forecast initialization time within the testing dataset ($\textrm{D}$), and ${\tau}$ is the forecast lead time.
Indices $i$ and $j$ denote latitude and longitude grid points, respectively, with $\textrm{H}$ and $\textrm{W}$ representing the grid dimensions.
The latitude-dependent weighting factor $\alpha_i= \textrm{H}\times\left.{cos\Phi_i}\middle/ \right.{\displaystyle\sum_{i=1}^\textrm{H}cos\Phi_i}$ accounts for the decreasing grid-cell area toward the poles.
The climatological mean ($\textrm{\textbf{M}}$) is computed from ERA5 over the period 1993-2016.


To quantitatively assess spatial downscaling performance from 1.5$^{\circ}$ to 0.25$^{\circ}$, we compare FuXi-Uni with bilinear interpolation using RMSE and the peak signal-to-noise ratio (PSNR).
PSNR is defined as the logarithmic ratio between the maximum possible signal power and the mean squared error (MSE) between the downscaled and reference fields, and is widely used to assess reconstruction fidelity in super-resolution and downscaling studies \cite{huynh2008scope,zhong2024investigating}.
Higher PSNR values indicate closer agreement with the reference data and better preservation of fine-scale spatial structures.

\subsubsection{Tropical cyclone track and intensity evaluation}

Tropical cyclones (TCs) in FuXi-Uni forecasts are identified and tracked using the Aurora tracker \cite{bodnar2025foundation}. Forecast skill is evaluated in terms of both track and intensity.
Track performance is quantified using the mean absolute error (MAE), defined as the distance between the forecast and observed TC center positions.
TC intensity is evaluated using the maximum $\textrm{WS10}$ near the TC center.
Intensity forecast accuracy is evaluated using the RMSE relative to IBTrACS..

\subsubsection{Evaluation metrics for biomedical VQA}
\label{evaluation_VQA}


For VQA-RAD, SLAKE, and PathVQA, results are commonly reported separately for \textbf{Open} and \textbf{Closed} questions \cite{VQA_RAD,SLAKE}.
In these benchmarks, Closed questions largely correspond to a closed-set setting (e.g., yes/no or fixed-choice) and are evaluated by accuracy with exact match between prediction and ground truth.
Open questions are closer to an open-set free-form setting; exact string match can be overly strict due to synonyms and minor wording variations.
Following recent medical LMM evaluation conventions (e.g., LLaVA-Med and MedTrinity-based settings), we evaluate Open questions using a relaxed keyword matching metric, i.e., token-level recall defined as the ratio of ground-truth tokens that appear in the generated sequence \cite{li2023llava,xie2024medtrinity}.
We note that some prior works formulate Open questions as classification by treating unique training answers as candidate labels; in contrast, we do not constrain the generated responses, which better reflects the open-set nature but is intrinsically more challenging.
This Open/Closed split therefore captures two difficulty regimes: Open emphasizes precise medical concept generation, whereas Closed emphasizes reliable clinical discrimination.

\section*{Data Availability}

We downloaded subsets of the 0.25$^\circ$ and 1.5$^\circ$ ERA5 datasets from the Copernicus Climate Data Store (CDS). The ERA5 data were obtained from \url{https://cds.climate.copernicus.eu/datasets}.
ECMWF HRES TC tracks were retrieved from the TIGGE archive in the form of downloadable XML files, which can be accessed via \url{https://confluence.ecmwf.int/display/TIGGE/Tools}. Additionally, we obtained the ground truth tracks of TC from the International Best Track Archive for Climate Stewardship (IBTrACS) project, which is publicly available at \url{https://www.ncei.noaa.gov/products/international-best-track-archive}.
We also used two publicly available medical vision-language datasets: LLaVA-Med (\url{https://github.com/microsoft/LLaVA-Med}) and MedTrinity-25M (\url{https://huggingface.co/datasets/UCSC-VLAA/MedTrinity-25M}).

\section*{Code Availability}

The Aurora TC tracker used in this study is available at \url{https://github.com/microsoft/aurora} \cite{bodnar2025foundation}.

\section*{Acknowledgments}

This work was supported by AI for Science Program, Shanghai Municipal Commission of Economy and Information.
We extend our sincere appreciation to the researchers at ECMWF for their invaluable contributions in collecting, archiving, disseminating, and maintaining the ERA5 reanalysis dataset and ECMWF-HRES.

The computations in this research were performed using the CFFF platform of Fudan University.

\section*{Competing interests}
The authors declare no competing interests.

\noindent


\bibliography{refs}


\end{CJK*}

\clearpage

\appendix

\renewcommand\thefigure{\thesection.\arabic{figure}}    

\setcounter{figure}{0}    
\setcounter{table}{0}    

\renewcommand{\figurename}{Extended Data Fig}
\renewcommand{\tablename}{Extended Data Table}

\end{document}